%% file: main.tex
\documentclass{article}
\usepackage{natbib} 
    \usepackage[final]{neurips_2023}

\usepackage[utf8]{inputenc} 
\usepackage[T1]{fontenc}    
\usepackage{hyperref}       
\usepackage{url}            
\usepackage{booktabs}       
\usepackage{amsfonts}       
\usepackage{nicefrac}       
\usepackage{microtype}      
\usepackage{xcolor}         
\input{settings}

\title{Approximate inference of marginals using the IBIA framework}

\author{%
  Shivani Bathla \\
  Department of Electrical Engineering\\
  Indian Institute of Technology Madras \\
  India, 600036 \\
  \texttt{ee13s064@ee.iitm.ac.in} \\
  \And
  Vinita Vasudevan \\
  Department of Electrical Engineering\\
  Indian Institute of Technology Madras \\
  India, 600036 \\
  \texttt{vinita@ee.iitm.ac.in} \\
}

\begin{document}

\maketitle

\input{abstract}
\input{introduction1}
\input{background}
\input{inferMAR}

\input{results.tex}
\input{conclusions.tex}
\bibliography{main}
\input{supplementary}

\end{document}

%% file: settings.tex
\usepackage[T1]{fontenc}
\usepackage{array}
\usepackage{lipsum,calc}
\usepackage{verbatim}
\usepackage{float}
\usepackage{multirow}
\usepackage{amsmath, amssymb}
\usepackage{amsthm}
\usepackage{tablefootnote}
\usepackage{pgfplots}
\pgfplotsset{compat=1.18} 
\usepackage{wrapfig,lipsum,booktabs}
\usepackage{threeparttable, tablefootnote}
\usepackage{colortbl}
\usepackage{xcolor}
\usepackage{hhline}
\usepackage{graphics}
\usepackage{bigstrut} 
\usepackage{soul}
\usepackage{url}
\usepackage{enumitem} 

\makeatletter
\let\MYcaption\@makecaption
\makeatother
\usepackage[font=footnotesize]{subcaption}
\makeatletter
\let\@makecaption\MYcaption
\makeatother

\makeatletter
\newcommand*{\rom}[1]{\expandafter\@slowromancap\romannumeral #1@}
\makeatother

\floatstyle{ruled}
\newfloat{algorithm}{tbp}{loa}
\providecommand{\algorithmname}{Algorithm}
\floatname{algorithm}{\protect\algorithmname}

\usepackage{algorithm, algpseudocode}

\algnewcommand{\Initialize}[1]{\State \textbf{Initialize: \newline}}
\algnewcommand{\LineComment}[1]{\State  \(\triangleright\) #1 \hfill~}
\algnewcommand{\IIf}[1]{\State\algorithmicif\ #1\ \algorithmicthen\ }
\algnewcommand{\IElse}{\unskip\ \algorithmicelse\ }
\algnewcommand{\EndIIf}{\unskip}
\algnewcommand{\pElse}[1]{\State \algorithmicelse\ #1}

\algnewcommand{\FFor}[1]{\State\algorithmicfor\ #1 \algorithmicdo }
\algnewcommand{\EndFFor}{\unskip}

\theoremstyle{definition}
\newtheorem{definition}{Definition}
\theoremstyle{proposition}
\newtheorem{proposition}{Proposition}
\theoremstyle{theorem}
\newtheorem{theorem}{Theorem}
\theoremstyle{corollary}
\newtheorem{corollary}{Corollary}
\newenvironment{hproof}{%
  \proof}{\endproof}

\newcounter{step}[section]

\usepackage{tikz}
\usepackage{pgfplots}
\usetikzlibrary{automata, positioning, shapes.geometric, arrows, decorations, fit, backgrounds}
\usetikzlibrary{trees,positioning,shapes,shadows,arrows}
\usetikzlibrary{arrows.meta}
\usepgfplotslibrary{statistics}
\usepgfplotslibrary{groupplots}
\usetikzlibrary{backgrounds}

\tikzset{orgClique/.style={thick, draw=gray, rectangle, align=center, minimum height=16pt}}
\tikzset{pathClique/.style={thick, draw=gray, rectangle, align=center, fill=red!20, minimum height=16pt}}
\tikzset{newSubtreeClique/.style={thick, draw=gray, rectangle, align=center, fill=teal!20, minimum height=16pt}}
\tikzset{newClique/.style={thick, draw=gray, rectangle, align=center, fill=teal!20, minimum height=16pt}}
\tikzstyle{loop1} = [draw,line width=5pt,-,red!20]
\tikzstyle{loop2} = [draw,line width=5pt,-,teal!20]	
\tikzstyle{loop3} = [draw,line width=5pt,-,magenta!40]
\tikzstyle{loopb} = [draw,line width=5pt,-,blue!20]
\tikzset{graphNode/.style={draw=gray, circle, align=center, inner sep=1pt, minimum size=10pt}}
\tikzset{reverseclip/.style={insert path={(-100cm,100cm) rectangle (100cm,-100cm)}}}
\definecolor{mygray}{rgb}{.906,  .902,  .902}
\definecolor{mygreen}{rgb}{.886,  .937,  .855}
\definecolor{myblue}{rgb}{.867,  .922,  .969}
\definecolor{myyellow}{rgb}{1,  .949,  .8}
\definecolor{myred}{rgb}{.969,  .835,  .859}

\usetikzlibrary{shapes.multipart}
\tikzset{
state/.style={
       rectangle split,
       rectangle split parts=2,
       rectangle split part fill={red!10,white},
       rounded corners,
       draw=black, thick,
       minimum height=4em,
       text width=3.5cm,
       inner sep=2pt,
       text centered,
       }
}

\usepackage{float}
\usepackage{caption}
\usepackage{pifont}

\usepackage{pgfplots}
\usepackage{tikz}
\usetikzlibrary{decorations.pathreplacing}

\tikzset{roundnode/.style={circle,draw, minimum size=3mm,inner sep=2pt},
  squarenode/.style={rectangle, draw=blue!60, very thick, minimum size=3mm}}

\tikzset{fatarrow/.style={single arrow,shape border rotate=270,
                          thick,draw=blue!70,fill=blue!30,
                          minimum height=10mm}}

\tikzstyle{vertex}=[circle,fill=black!20,minimum size=10pt,inner sep=1pt,draw=black!50]
\tikzstyle{vertexpt}=[circle,fill=black!25,minimum size=10pt,inner sep=0pt]
\tikzstyle{vertexr}=[rectangle,fill=black!15,minimum size=20pt,inner sep=2pt]
\tikzstyle{vertexr1}=[rectangle,fill=black!15,minimum size=20pt,inner sep=1pt]
\tikzstyle{vertexrnofill}=[rectangle,draw=black,minimum size=20pt,inner sep=2pt]
\tikzstyle{str}=[rectangle,fill=black!15,draw=black,minimum size=20pt,inner sep=2pt]
\tikzstyle{selected vertex}=[rectangle,fill=red!15,minimum size=20pt,inner sep=2pt]
\tikzstyle{selected vertext}=[rectangle,fill=teal!20,minimum size=20pt,inner sep=2pt]
\tikzstyle{selected vertexy}=[rectangle,fill=orange!20,minimum size=20pt,inner sep=2pt]
\tikzstyle{selected vertexr}=[rectangle,fill=cyan!15,minimum size=20pt,inner sep=2pt]
\tikzstyle{selected vertexb}=[rectangle,fill=blue!15,minimum size=20pt,inner sep=2pt]
\tikzstyle{edge} = [draw,thick,->]
\tikzstyle{uedge} = [draw,thick]
\tikzstyle{dashededge} = [draw,dashed,thick,-]
\tikzstyle{udashededge} = [draw,ultra thick,dashed,magenta]
\tikzstyle{tealedge} = [draw,ultra thick,->,teal]
\tikzstyle{blackedge} = [draw,ultra thick,->]
\tikzstyle{edge1} = [draw,thick,<-]
\tikzstyle{edge2} = [draw,thick,-]
\tikzstyle{edge3} = [draw,thick,blue,->]
\tikzstyle{edge4} = [draw,blue,ultra thick]
\tikzstyle{edge5} = [->, >=latex, line width=1pt]
\tikzstyle{weight} = [font=\small,blue]
\tikzstyle{selected edge} = [draw,line width=5pt,->,red!50]
\tikzstyle{ignored edge} = [draw,line width=5pt,-,black!20]
\tikzstyle{loop1} = [draw,line width=5pt,-,red!20]
\tikzstyle{loop2} = [draw,line width=5pt,-,teal!20]	
\tikzstyle{loop3} = [draw,line width=5pt,-,magenta!40]
\usetikzlibrary{arrows}
\usetikzlibrary{shapes.geometric,backgrounds}

%% file: abstract.tex
\begin{abstract}
    Exact inference of marginals in probabilistic graphical models (PGM) is known to be intractable, necessitating the use of approximate methods. Most of the existing variational techniques perform iterative message passing in loopy graphs which is slow to converge for many benchmarks. In this paper, we propose a new algorithm for marginal inference that is based on the incremental build-infer-approximate (IBIA) paradigm. Our algorithm converts the PGM into a \textit{sequence of linked clique tree forests (SLCTF)} with bounded clique sizes, and then uses a heuristic belief update algorithm to infer the marginals. For the special case of Bayesian networks, we show that if the incremental build step in IBIA uses the topological order of variables then (a) the prior marginals are consistent in all CTFs in the SLCTF  and (b) the posterior marginals are consistent once all evidence variables are added to the SLCTF. In our approach, the belief propagation step is non-iterative and the accuracy-complexity trade-off is controlled using user-defined clique size bounds. Results for several benchmark sets from recent UAI competitions show that our method gives either better or comparable accuracy than existing variational and sampling based methods, with smaller runtimes.
\end{abstract}

%% file: introduction1.tex
\section{Introduction}\label{sec:intro}
Discrete probabilistic graphical models (PGM) including Bayesian networks (BN) and Markov networks (MN) are used for probabilistic inference in a wide variety of applications.
An important task in probabilistic reasoning is the computation of posterior marginals of all the variables in the network.
Exact inference is known to be \#P-complete~\citep{Roth1996}, thus necessitating approximations. Approximate techniques can be broadly classified as sampling based and variational methods.

Sampling based methods include Markov chain Monte Carlo based techniques like Gibbs sampling~\citep{Gelfand2000,Kelly2019} and importance sampling based methods~\citep{Gogate2011,Friedman2018,Kask2020,Broka2018,Lou2019,Lou2017search,Lou2017DIS,Marinescu2019,Marinescu2018}.
An advantage of these methods is that accuracy can be improved with time without increasing the required memory. However, in many benchmarks the improvement becomes slow with time. 
\textcolor{black}{Moreover, many of the recent sampling/search based techniques \citet{Kask2020,Broka2018,Lou2019,Lou2017search,Lou2017DIS,Marinescu2019,Marinescu2018} have been evaluated either for approximate inference of partition function (PR) or for finding the marginal maximum a posteriori assignment (MMAP). Currently, there are no published results for posterior marginals (MAR) using these methods, and the publicly available implementations do not support the MAR task.}
Alternatively, variational techniques can be used.  
These include loopy belief propagation (LBP)~\citep{Frey1998}
region-graph based techniques like generalized belief propagation (GBP)~\citep{Yedidia2000} and its variants~\citep{HAK2003,Mooij07,Lin2020},
mini-bucket based schemes like iterative join graph propagation (IJGP)~\citep{Mateescu2010} and weighted mini-bucket elimination (WMB)~\citep{Liu2011} and methods that simplify the graph structure like edge deletion belief propagation (EDBP) and the related relax-compensate-recover (RCR) techniques~\citep{Choi05,Choi2006,Choi2010}.
While the accuracy-complexity trade-off can be achieved using a single user-defined clique size bound in mini-bucket based methods, it is non-trivial in many of the other region graph based methods.
Most of these techniques use iterative message passing to solve an optimization problem, for which convergence is not guaranteed and even if possible, can be slow to achieve. \textcolor{black}{Non-iterative methods like Deep Bucket Elimination (DBE)~\citep{DBE} and NeuroBE~\citep{NeuroBE} are extensions of bucket elimination that approximate messages using neural networks. However, training these networks takes several hours. 
Moreover, the publicly available implementations of these methods do not support the MAR task.
}

The recently proposed \textit{incremental build-infer-approximate} (IBIA) framework~\citep{IBIAPR} uses a different approach. It converts the PGM into a sequence of calibrated clique tree forests (SCTF) with clique sizes bounded to a user-defined value. 
\citet{IBIAPR} show that the normalization constant (NC) of clique beliefs in the last CTF in the sequence is \textcolor{black}{a good approximation} of the partition function of the overall distribution. 
This framework has two main advantages. Firstly, since it is based on clique trees and not loopy graphs, the belief propagation step is non-iterative. Therefore, it is fast and has no issues related to convergence. Secondly, it provides an easy control of the accuracy complexity trade-off using two user-defined parameters \textcolor{black}{  and hence can be used in anytime manner.} 
However, the framework in~\citet{IBIAPR} cannot be used to infer marginals. 
This is because only the clique beliefs in the last CTF account for all factors in the PGM.
Beliefs in all other CTFs account for a subset of factors and thus, cannot be used for inference of marginals.

\textbf{Contributions of this work:} 
In this paper, we propose a method for marginal inference that uses the IBIA framework. 
We show that the approximation algorithm used in this framework preserves the within-clique beliefs. Based on this property, we modify the data structure generated by IBIA to add links between adjacent CTFs. 
We refer to the modified data structure as a \textit{sequence of linked clique tree forests} (SLCTF).
We propose a heuristic belief update algorithm 
\textcolor{black}{that back-propagates beliefs from the last CTF to the previous CTFs via the links and re-calibrates each CTF so that the updated beliefs account for all factors in the PGM. We also propose a greedy heuristic for the choice of links used for belief update.}
Results for several UAI benchmark sets show that our method gives 
\textcolor{black}{an accuracy that is better than or comparable}
to the existing variational and sampling based methods, with competitive runtimes.

For the special case of BNs, we show that if 
the incremental build step in IBIA is performed in the topological order of variables
then (a) the estimated partition function is guaranteed to be one if no evidence variables are present (b) the prior marginals of all variables are consistent across all CTFs in the sequence 
and  (c) once all the evidence variables have been added to the SLCTF, the posterior marginals of variables in subsequent CTFs are consistent.
Our results show that using the topological ordering for BNs leads to better estimates of partition function, prior marginals and posterior marginals in most benchmarks.

%% file: background.tex
\section{Background}
\subsection{Discrete Probabilistic Graphical Models}
Let  $\mathcal{X} = \{ X_1, X_2, \cdots X_n\}$ be a set of random variables with associated domains $D = \{D_{X_1},D_{X_2}, \cdots D_{X_n}\}$. The probabilistic graphical model (PGM) over $\mathcal{X}$ consists of a set of factors, $\Phi$, where each factor $\phi_{\alpha}(\mathcal{X}_\alpha) \in {\Phi}$ is defined over a subset of variables, $Scope(\phi_\alpha)=\mathcal{X}_\alpha$. If $D_\alpha$ denotes the Cartesian product of the domains of variables in $\mathcal{X}_\alpha$, then $\phi_{\alpha}: D_\alpha \rightarrow R \geq 0$. The joint probability distribution captured by the PGM is $P(\mathcal{X})=\frac{1}{Z}\prod_{\alpha}\phi_{\alpha}$, where $Z=\sum_{\mathcal{X}} \prod_{\alpha}\phi_{\alpha}$ is the partition function. 
PGMs can be broadly classified as Markov networks (MN) which are the undirected models and Bayesian networks (BN) which are the directed models.


One method to perform exact inference involves converting the PGM into a \textit{clique tree} (CT), which is a hypertree where each node $C_i$ is a clique that contains a subset of variables. 
We use the term $C_i$ as a label for the clique as well as to denote the set of variables in the clique. 
An edge between $C_i$ and $C_j$ is associated with a set of \textit{sepset variables} denoted $S_{i,j} = C_i \cap C_j$. 
Exact inference in a CT is done using the belief propagation (BP) algorithm~\citep{Lauritzen1988} that is equivalent to two rounds of message passing along the edges of the CT. The algorithm returns a CT with calibrated clique beliefs $\beta(C_i)$ and sepset beliefs $\mu(S_{i,j})$. 
In a calibrated CT, all clique beliefs have the same normalization constant ($Z$) and beliefs of all adjacent cliques agree over the marginals of the sepset variables.
The joint probability distribution, $P(\mathcal{X})$,  can be rewritten as follows.
\begin{equation}\label{eq:reparam}
  P(\mathcal{X}) = \frac{1}{Z}\frac{\prod_{i \in \mathcal{V}_T} \beta(C_i)}{\prod_{(i,j)\in \mathcal{E}_T}\mu(S_{i,j})}
\end{equation}
where $\mathcal{V}_T$ and $\mathcal{E}_T$ are the set of nodes and edges in the CT.
The marginal probability distribution of a variable $X_i$ can be obtained by marginalizing the belief of any clique $C$ that contains $X_i$ as follows.
\begin{equation}\label{eq:marg}
    P(X_i)=\frac{1}{Z}\sum\limits_{C\setminus X_i}\beta(C)
\end{equation}
We use the following definitions in this paper.
\begin{definition}
Clique Tree Forest (CTF): Set of disjoint clique trees.
\end{definition}
\begin{definition}\label{def:cs}
Clique size:
The clique size $cs$ of a clique $C$ is the effective number of binary variables contained in $C$. It is computed as follows.
  \begin{equation}\label{eqn:cs}
    cs = \log_2~(\prod\limits_{X_i~\in~ C} |D_{X_i}| ~)
  \end{equation}
    where $|D_{X_i}|$ is the cardinality or the number of states in the domain of the variable $X_i$.
\end{definition}

\begin{definition}
    Prior marginals ($P(X_i)$): It is the marginal probability of a variable $X_i$ when the PGM has no evidence variables.
\end{definition}
\begin{definition}
    Posterior marginals ($P(X_i|E=e)$): It is the conditional probability distribution of a variable $X_i$, given a fixed evidence state $e$.
\end{definition}

\subsection{Overview of IBIA framework}\label{sec:overview}
\input{tikz/ibiaExample2}
\textbf{Methodology:}
The IBIA framework proposed in~\citet{IBIAPR} converts the PGM into a sequence of calibrated CTFs (SCTF) with bounded clique sizes.
IBIA starts with an initial CTF, $CTF_0$, that contains cliques corresponding to factors with disjoint scope.
Figure~\ref{fig:ibiaFramework} illustrates the overall methodology used in the framework. 
It uses three main steps as described below.

\textit{Incremental Build:} In this step, a CTF, $CTF_k$, is constructed by incrementally adding factors in the PGM to an initial CTF ($CTF_0$ or $CTF_{k-1,a}$) until the maximum clique size reaches a user-specified bound $mcs_p$.
Methods for incremental construction of CTs have been proposed in~\citet{IBIAPR} and ~\citet{Flores2002}. Either of the two methods can be used to incrementally add new factors.

\textit{Infer:} In this step, all CTs in $CTF_k$ are calibrated using the standard belief propagation algorithm. 

\textit{Approximate:} In this step, $CTF_k$ is approximated to give an approximate CTF, $CTF_{k,a}$, that has clique sizes reduced to another user-specified bound $mcs_{im}$. 

As shown in Figure~\ref{fig:ibiaFramework}, IBIA constructs the $SCTF=\{CTF_1,\hdots,CTF_n\}$ by repeatedly using the incremental build, infer and approximate steps. This process continues until all factors in the PGM are added to some CTF in the SCTF. 
Figure~\ref{fig:ibiaExample} shows the SCTF generated by IBIA for an example PGM, $\Phi$ (specified in the caption to the figure), with clique size bounds $mcs_p$ and $mcs_{im}$  set to 4 and 3 respectively. 
For the purpose of illustrating all steps, the disjoint cliques corresponding to factors $\phi(d,m,o),\phi(i,l)$ and $\phi(j,h)$ are chosen as the initial CTF, $CTF_0$.
$CTF_1$ is constructed by incrementally adding factors to $CTF_0$. 
All factors except $\phi(k,l,o)$ and $\phi(f,o)$ are added to $CTF_1$. These two factors are deferred since their addition results in clique sizes greater than $mcs_p=4$.
$CTF_1$ is calibrated using BP and then approximated to give $CTF_{1,a}$ with clique sizes bounded to $mcs_{im}=3$. $CTF_2$ is constructed by adding the remaining factors to $CTF_{1,a}$. We will use this example to explain the steps in our method for inference of marginals.
\textbf{Approximate step:}
Since our inference algorithm is based on the properties of the approximate CTF, we explain this step in more detail using the running example shown in Figure~\ref{fig:ibiaExample}.
Variables $f,k,l$ and $o$ in $CTF_1$ are also present in the deferred factors $\phi(k,l,o)$ and $\phi(f,o)$. These variables are needed for the construction of subsequent CTFs and are called \textit{interface variables} (IV). All other variables in $CTF_1$ are called \textit{non-interface variables} (NIV). $CTF_{1,a}$ is initialized as the minimal subgraph that connects the IVs. This subgraph contains all cliques in $CTF_1$ except clique $fmn$. 
Approximation involves two main steps to reduce the number of cliques and clique sizes. \\
1. \textit{Exact marginalization:} 
The goal of this step is to remove as many NIVs as possible while ensuring that the overall joint distribution is preserved.
NIV $j$ is present in a single clique $hmj$ and is marginalized out from it by summing over the states of $j$. The resulting clique $hm$ is a non-maximal clique that is contained in clique $fdhm$, and is thus removed.
NIV $i$ is removed after collapsing \textcolor{black}{the two} containing cliques $hi$ and $il$. 
Exact marginalization of the other NIVs results in collapsed cliques with size greater than $mcs_{im}=3$, and is not performed. \\
2. \textit{Local marginalization:} In this step, clique sizes are reduced by marginalizing variables from individual cliques with size greater than $mcs_{im}$ while ensuring that (a) $CTF_{1,a}$ is a valid CTF that satisfies the running intersection property (RIP) (b) a connected CT in $CTF_1$ remains connected in $CTF_{1,a}$ and (c) $CTF_{1,a}$ contains all IVs. 
To reduce the size of the large-sized clique $fdhm$, NIV $d$ is locally marginalized from this clique. In order to satisfy RIP, it needs to be marginalized from either clique $dmo$ or $dhk$ \textcolor{black}{as well as the corresponding sepsets}. Here, $d$ is locally marginalized from clique $dmo$ \textcolor{black}{and the corresponding sepsets} to give $CTF_{1,a}$ as shown in the figure. 
Since all cliques containing $d$ are not collapsed before marginalization, this results in an approximate joint distribution. 
\\
\textcolor{black}{Propositions~5 and 6 in }\citet{IBIAPR} show that all CTs in the approximate CTF obtained after exact and local marginalization are valid and calibrated.

%% file: tikz/ibiaExample2.tex
\begin{figure}[t]
    \begin{subfigure}[t]{0.4\textwidth}
\begin{tikzpicture}[thick,scale=0.8, every node/.style={scale=0.8}]
\node (ib) [state,text width=2cm] at (0,0) {\textbf{Incremental Build}\nodepart{two} Add factors while max-clique-size $\leq mcs_p$};
\draw [edge5] (ib.east)--++(1.5,0) node [midway, above] {$\mathbf{CTF_k}$};
\draw [edge5] ($(ib.west)+(-1,0)$)--(ib.west) node [midway, above] {$\mathbf{\Phi}$}node [align=left,midway, below] {$k=1$\\$CTF_0$};
\node (inf) [state,text width=2cm, rectangle split part fill={blue!10,white}] at (3.5,0) {\textbf{Infer\vspace{7pt}}\nodepart{two} Calibrate using Belief Propagation\vspace{7pt}};
\node (check) [draw, diamond, aspect=1.5, inner sep=2pt, fill=yellow!10] at (3.5,-4) {$\Phi.empty()$};
\node (app) [state,text width=2cm, rectangle split part fill={teal!30,white}] at (0,-4) {\textbf{Approximate\vspace{7pt}}\nodepart{two} Reduce clique sizes to $mcs_{im}$\vspace{7pt}};
\draw [edge5] (app) -- (ib) node [align=left,midway,right] {$\mathbf{CTF_{k,a}}$\\ $\mathbf{k =k+1}$};
\draw [edge5] (check) -- (app) node [midway,above] {\textbf{No}};
\draw [edge5] (inf) -- (check) node [midway,right] {$\mathbf{CTF_k}$};
\node (stop) [draw,rectangle,rounded corners, below=0.75 of check] {\textbf{Stop}};
\draw [edge5] (check) -- (stop) node [right, midway] {\textbf{Yes}};
\end{tikzpicture}
\caption{}
        \label{fig:ibiaFramework}
\end{subfigure}
\hfill 
    \begin{subfigure}[t]{0.58\textwidth}
\begin{tikzpicture}[thick,scale=0.8, every node/.style={scale=0.8}]
    \node (phi) [align=center] at (1,-0.5) {$\mathbf{\Phi}$};
    \node (ctf0) at (3.5,-0.5) {$\mathbf{CTF_0}$};
    \node (ctf1) at (7.5,-0.5) {$\mathbf{CTF_1}$};
    \node (ctf1a) at (7.5,-5.8) {$\mathbf{CTF_{1,a}}$};
    \node (ctf2) at (2.5,-5.8) {$\mathbf{CTF_2}$};
    \draw [edge5] (phi) -- (ctf0) node[above,midway] {Initialize};
    \draw [edge5] (ctf0) -- (ctf1) node[align=left, above, midway] {Incremental Build} node [below, midway] {+ Infer};
    \draw [edge5] ($(ctf1a)+(0,1.5)$) -- (ctf1a) node[align=left, right, midway] {Approximate};
    \draw[edge5] (ctf1a) -- (ctf2) node [align=left,above,midway] {Incremental Build} node [below, midway] {+ Infer};

    \def\startx{2.25}
    \def\starty{-1.3}
    \def\xdistcliq{1.5}
    \def\ydistcliq{1.25}
    \foreach \xpos/\ypos/\name in {0/0/dmo, 0.75/0/jh, 1.5/0/il}
    \node (\name) [vertexr] at (\startx+\xpos*\xdistcliq, \starty+\ypos*\ydistcliq)  {$\name$};

    \def\startx{6}
    \def\starty{-1.3}
    \def\xdistcliq{1.5}
    \def\ydistcliq{1.25}
    \foreach \xpos/\ypos/\name in {2/-2/dmo, 1/-1/fdhm, 2/-1/hmj, 1/-2/dhk, 2/-1/hmj, 2/0/hi, 2.7/0/il, 1/0/fmn}
    \node (\name) [vertexr1,fill=blue!10] at (\startx+\xpos*\xdistcliq, \starty+\ypos*\ydistcliq)  {$\name$};
    
    \foreach \source/\dest/\weight  in { fdhm/hmj/$hm$, hi/il/$i$}
    \path[edge2] (\source) -- node [above,midway] {\color{gray} \weight} (\dest);
    \foreach \source/\dest/\weight  in { hmj/hi/$h$,dmo/fdhm/$dm$}
    \path[edge2] (\source) -- node [right,midway] {\color{gray} \weight} (\dest);
    \foreach \source/\dest/\weight  in {dhk/fdhm/$dh$,fmn/fdhm/$fm$}
    \path[edge2] (\source) -- node [left,midway] {\color{gray} \weight} (\dest);

    \def\starty{-6.5}
    \def\startx{7}
    \foreach \xpos/\ypos/\name in {1/-1/mo, 0/0/fhm, 1/0/hl, 0/-1/dhk}
    \node (\name) [vertexr1] at (\startx+\xpos*\xdistcliq, \starty+\ypos*\ydistcliq)  {$\name$};
    
    \foreach \source/\dest/\weight  in {fhm/hl/$h$}
    \path[edge2] (\source) -- node [above,midway] {\color{gray} \weight} (\dest);
    \foreach \source/\dest/\weight  in {mo/fhm/$m$,dhk/fhm/$h$}
    \path[edge2] (\source) -- node [right,midway] {\color{gray} \weight} (\dest);

    \def\starty{-6.5}
    \def\startx{1}
    \foreach \xpos/\ypos/\name in {0/0/fmoh, 1/0/ohkl, 2/0/dhk}
    \node (\name) [vertexr1,fill=blue!10] at (\startx+\xpos*\xdistcliq, \starty+\ypos*\ydistcliq)  {$\name$};

    \foreach \source/\dest/\weight  in {fmoh/ohkl/$ho$,ohkl/dhk/$hk$}
    \path[edge2] (\source) -- node [above,midway ] {\color{gray} \weight} (\dest);
\end{tikzpicture}
        \caption{}
        \label{fig:ibiaExample}
\end{subfigure}
\caption{Conversion of the PGM, $\Phi$, into a sequence of calibrated CTFs (SCTF) using the IBIA framework. (a) Overall methodology. (b) Construction of $SCTF=\{CTF_1,CTF_2\}$ for a PGM, $\Phi=\{\phi(d,m,o),\phi(j,h),\phi(i,l),\phi(f,m,n), \phi(h,i), \phi(d,h,k), \phi(f,d,h),\phi(j,m),\phi(k,l,o),\phi(f,o)\}$, with $mcs_p$ and $mcs_{im}$ set to 4 and 3 respectively. 
$CTF_0$ is formed using cliques corresponding to factors $\phi(d,m,o),\phi(j,h),\phi(i,l)$. $CTF_1$ contains all factors except $\phi(k,l,o),\phi(f,o)$. These factors are added to $CTF_2$.
}
\end{figure}
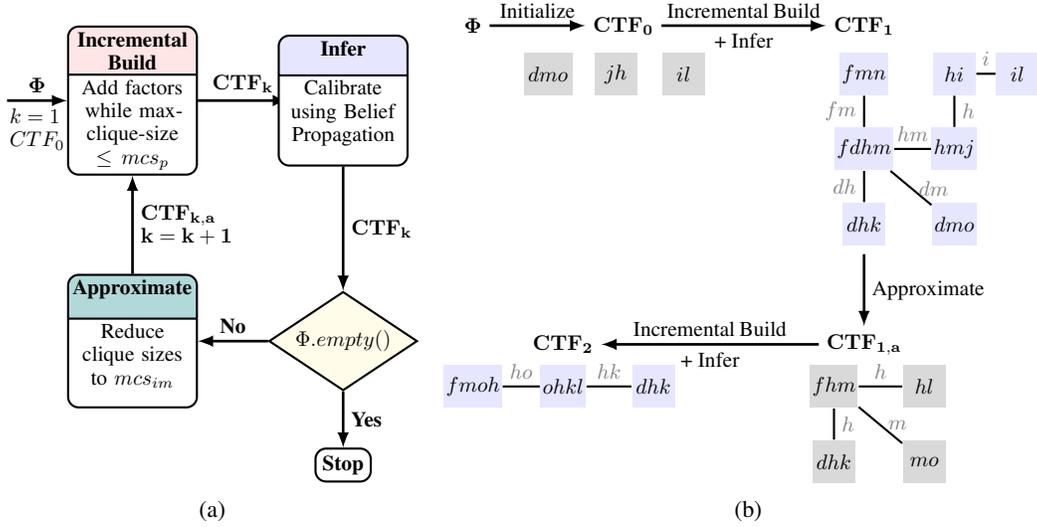

%% file: inferMAR.tex
\section{Inference of marginals}
In this section, we first discuss some of the properties satisfied by the SCTF generated by IBIA. Based on these properties, we then describe the proposed methodology for inference of marginals. 

\subsection{Properties of SCTF}\label{sec:properties}
We show that each $CTF_k$ in  SCTF and the corresponding approximate CTF, $CTF_{k,a}$, satisfy the following properties.
The detailed proofs for these properties are included in the supplementary material and the main ideas used in the proofs are discussed here.
\begin{proposition}\label{prop:appBeliefs}
    The joint belief of variables contained within any clique in the approximate CTF, $CTF_{k,a}$, is the same as that in $CTF_k$.
\end{proposition}
\begin{hproof}
    Both exact and local marginalization involve summing clique beliefs over the states of a variable, which does not alter the joint belief of the remaining variables in the clique.
\end{hproof}


\begin{proposition}\label{prop:ctfkBeliefs}
    The clique beliefs in $CTF_k$ account for all factors added to \{$CTF_1, \hdots,CTF_k$\}.
\end{proposition}
\begin{hproof}
\textcolor{black}{$CTF_1$ is exact, with clique beliefs corresponding to the partial set of factors used to form $CTF_1$. 
$CTF_{1,a}$ is a calibrated CTF that is obtained after approximating $CTF_1$. Thus, the joint belief of variables in $CTF_{1,a}$ is an approximation of the beliefs encoded by factors added to $CTF_1$.
}
    $CTF_2$ is obtained after adding new factors to $CTF_{1,a}$. 
    Therefore, after calibration, clique beliefs in $CTF_2$ account for factors added to $CTF_1$ and $CTF_2$. 
    The proposition follows, since the same argument holds true for all CTFs in the sequence. 
\end{hproof}

BNs can be handled in a similar manner as MNs by using the undirected moralized graph corresponding to the BN. 
Another possibility is to use the directed acyclic graph (DAG) corresponding to the BN to guide the incremental build step. 
In contrast to MNs where all factors are un-normalized, each factor in a BN is the conditional probability distribution (CPD) of a variable $y$ given the state of its parents in the DAG ($Pa_y$).
Factors corresponding to the evidence variables are simplified based on the given state and hence become un-normalized.

For BNs, the following properties hold true if each CTF in the SCTF is built by adding factors in the topological order. By this, we mean that the factor corresponding to the variable $y$ is added only after the factors corresponding to all its parent variables, $Pa_y$, have been added to some CTF in the sequence.
Let $Y_k$ denote the set of variables whose CPDs are added during construction of $CTF_k$, ~~$e_k$ denote the evidence states of all evidence variables in $Y_k$ and $Pa_{Y_k}$ denote the parents of variables in the set $Y_k$.

\begin{proposition}\label{prop:topo}
The product of factors added in CTFs, $\{CTF_1,\hdots,CTF_k\}$ is a valid \textcolor{black}{joint} probability distribution whose normalization constant is the probability of evidence states $e_1,\hdots,e_k$. 
\end{proposition}
\begin{proof}
    For each variable $y\in\{Y_1,\hdots,Y_k\}$, the corresponding CPD, $P(y|Pa_y)$, is added to some CTF in $\{CTF_1,\hdots,CTF_k\}$.
    The proposition follows since the CPDs corresponding to parents of $y$ are always added to a CTF before the CPD of $y$ is added. 
\end{proof}
%
%

\begin{proposition}\label{prop:NC}
    The normalization constant of the distribution encoded by the calibrated beliefs in $CTF_k$ is the estimate of probability of evidence states $e_1, \hdots, e_k$.
\end{proposition}
\begin{hproof}
    Using Proposition~\ref{prop:topo}, the normalization constant (NC) of the distribution encoded by $CTF_1$ is $P(e_1)$.
    Using Proposition~\ref{prop:appBeliefs}, \textcolor{black}{the approximation algorithm preserves the within-clique beliefs and hence the NC.}
    Thus, the NC of $CTF_{1,a}$ is also $P(e_1)$. Although the NC is the same, the overall distribution corresponding to $CTF_{1,a}$ is approximate due to local marginalization.  $CTF_2$ is constructed by adding CPDs of variables in $Y_2$ to $CTF_{1,a}$.
    CPDs of parent variables in $Pa_{Y_2}$ are added either in $CTF_1$ or $CTF_2$. 
    Hence, after calibration, we get a valid probability distribution with NC as the estimate of probability of evidence states $e_1,e_2$. 
    A similar procedure can be used to show that the property holds for all CTFs.
\end{hproof}

\begin{corollary}\label{prop:bnpr}
    For a BN with no evidence variables, the normalization constant of any CT in $CTF_k$ is guaranteed to be one.
\end{corollary}

\begin{theorem} \label{thm:post}
Let $I_E$ denote the index of the last CTF in the sequence where the factor corresponding to an evidence variable is added.
  The posterior marginals of variables present in CTFs $\{CTF_{k}, k \geq I_E\}$ are preserved and can be computed from any of these CTFs.
\end{theorem}
\begin{hproof}
    Once all evidence variables are added, additional CPDs added in each new CTF in $\{CTF_{k}, k> I_E\}$ correspond to the successors in the BN. \textcolor{black}{Since none of the successors are evidence variables, the corresponding CPDs are normalized and hence cannot change the beliefs of the previous variables.}
\end{hproof}
\begin{corollary} \label{thm:prior}
    For a BN with no evidence variables, the estimate of prior marginals obtained from any CTF in the sequence is the same.
\end{corollary}


\subsection{Proposed algorithm for inference of marginals}
We first explain our approach for estimation of marginals with the help of the example shown in Figure~\ref{fig:ibiaExample}. Following this, we formally describe the main steps in our algorithm. 

The SCTF generated by IBIA for the example in Figure~\ref{fig:ibiaExample} contains two CTFs, $CTF_1$ and $CTF_2$. 
Using Proposition~\ref{prop:ctfkBeliefs}, we know that calibrated clique beliefs in $CTF_2$ account for all factors in the PGM, $\Phi$.
Therefore, the marginals of all variables present in it can be inferred using Equation~\ref{eq:marg}. However, clique beliefs in $CTF_1$ do not account for factors $\phi(k,l,o)$ and $\phi(f,o)$ which were added during the construction of $CTF_2$. Therefore, in order to infer the marginals of variables $n,j,i$ that are present only in $CTF_1$, we need to update the beliefs to account for these two factors. 

Figure~\ref{fig:links} shows $CTF_1$, $CTF_{1,a}$ and $CTF_2$ for the example.
Using Proposition~\ref{prop:appBeliefs},
we know that the joint belief of variables present within any clique in $CTF_{1,a}$ is the same as that in $CTF_1$. 
However, this belief changes when new factors are added during the construction of $CTF_2$. 
For instance, $\beta({C_2}')=\sum_{d} \beta(C_2) \neq \sum_{o} \beta(\tilde{C_2})$.
To reflect the effect of new factors added in $CTF_2$, the joint belief of variables in clique $C_2$ can be updated as follows.
\begin{equation*} 
    \beta_{updated}(C_2) = \frac{\beta(C_2)}{\sum\limits_{d}\beta(C_2)}\sum_{o}\beta(\tilde{C_2})
\end{equation*}

To make sure that $CTF_1$ remains calibrated, this must be followed by a single round of message passing in $CTF_1$ with $C_2$ as the root node. It is clear that a similar belief update is needed for all the variables in $CTF_{1,a}$.
However, every update and subsequent round of message passing will override the previous updates. 
\input{tikz/links}
Hence, the beliefs in $CTF_1$ will only approximately reflect the effect of additional factors in $CTF_2$. To improve accuracy, we propose a heuristic procedure for belief update sequence.

Formally, the steps in our algorithm are as follows. 
Variables present in $CTF_{k,a}$ are present in both $CTF_k$ and $CTF_{k+1}$. We refer to these variables as the \textit{link variables}. 
We first find links between corresponding cliques $C\in CTF_k$, $C'\in CTF_{k,a}$ and $\tilde{C}\in CTF_{k+1}$. 
Each link $(C,C',\tilde{C})$ is associated with a set of link variables $V_l=C\cap C'$.
For the example, links between $CTF_1$, $CTF_{1,a}$ and $CTF_2$, and the corresponding link variables are shown in magenta in Figure~\ref{fig:links}.
The first part of a link contains cliques $C'$ and $C$. It is obtained as follows.

        (a)       If $C'$ is obtained after collapsing a set of cliques $\{C_1, \cdots C_m\}$ in $CTF_{k}$,  $C'$ is linked to each of $\{C_1, \cdots C_m\}$. 
For example, ${C_4}'$ is linked to $C_4$ and $C_5$, which were collapsed during exact marginalization of variable $i$.

        (b)    If $C'$ is obtained from $C$ in $CTF_{k}$ after local marginalization, $C'$ is linked to $C$. 
        In the example, cliques ${C_1}'$ and ${C_2}'$ are obtained after local marginalization of variable $d$ from cliques $C_1$ and $C_2$ respectively. Hence,  the corresponding tuples are $(C_1, {C_1}')$ and $(C_2,{C_2}')$.

        (c) If $C'$ is same as clique $C$ in $CTF_{k}$, $C'$ is linked to $C$. 
For example, $C_3$ is linked to ${C_3}'$.

The second part links $C'$ to $\tilde{C}$ in $CTF_{k+1}$ such that $C'\subseteq \tilde{C}$ . This is always possible since $CTF_{k+1}$ is obtained after incrementally modifying $CTF_{k,a}$ to add new factors. Thus, each clique in $CTF_{k,a}$ is contained in some clique in $CTF_{k+1}$.  
For example, ${C_1}'\subset \tilde{C_2}$ and the link is $(C_1, {C_1}',\tilde{C_2})$.

\textcolor{black}{Let $L_k$ denote the set of links between $CTF_k$ and $CTF_{k+1}$.}
We refer to the modified data structure consisting of the sequence of calibrated CTFs, $SCTF=\{CTF_k\}$ and a list of links between all adjacent CTFs, $SL=\{L_k\}$, as the \textit{sequence of linked CTFs (SLCTF)}.
Once the SLCTF is created, starting from the final CTF in the SLCTF, we successively back-propagate beliefs to the preceding CTFs via links between adjacent CTFs. 
To back-propagate beliefs from $CTF_{k+1}$ to $CTF_k$, we choose a subset of links in $L_k$ based on heuristics, which will be discussed later. Then, for each selected link $(C,C',\tilde{C})$, 
we update the belief associated with clique $C$ as follows.
\begin{align} \label{eq:bu}
    \beta_{updated}(C) &= \left( \frac{\beta(C)}{\sum\limits_{C\setminus V_l}\beta(C)} \right)  \sum \limits_{{\tilde{C}} \setminus {V_l}} \beta(\tilde{C})
\end{align}
\textcolor{black}{where, $V_l=C\cap C'$.}
This is followed by one round of message passing from $C$ to all other cliques in the CT containing $C$.
Once all CTFs are re-calibrated, we infer the marginal distribution of a variable 
using Equation~\ref{eq:marg} 
from the
last CTF in which it is present.

For BNs, if incremental build is performed by adding variables in the topological order, then as shown in Theorem~\ref{thm:post}, the singleton marginals are consistent in CTFs $\{CTF_{k\geq I_E}\}$, where $I_E$ is the index of the last CTF to which the factor corresponding to an evidence variable is added. Therefore, in this case, back-propagation of beliefs can be performed starting from $CTF_{I_E}$ instead of starting from the last CTF. This reduces the effort required for belief update.

\textbf{Heuristics for choice of links}:
To ensure that beliefs of all CTs in $CTF_{k}$ are updated, at least one link must be chosen for each CT.
It is also clear that for any CT in $CTF_k$ more than one link may be required since variables that have low correlations in $CTF_{k}$ could become tightly correlated when new factors are added in $CTF_{k+1}$.
 \textcolor{black}{However, belief update via all links is expensive since a round of message passing is required for each link}.
Based on results over many benchmarks, \textcolor{black}{we propose the following greedy heuristic} to choose and schedule links for backward belief update. 

   \textcolor{black}{(a)  To minimize the number of selected links, we first choose a subset of link variables 
        for which the difference in posterior marginals in adjacent CTFs, $CTF_k$ and $CTF_{k+1}$, is greater than a threshold. 
        Next, for belief update, we select a minimal set of links that cover these variables.}


        
        
        (b)  The updated beliefs depend on the order in which the links are used for update. Based on the difference in marginals, we form a priority queue with the cliques containing link variables that have the lowest change in marginals having the highest priority. This is to make sure that large belief updates do not get over-written by smaller ones. 
        This could happen for example, if two variables, $v_1$ and $v_2$, that are highly correlated in $CTF_k$ become relatively uncorrelated in $CTF_{k+1}$ due to the approximation. Assume that new factors added to  $CTF_{k+1}$ affect $v_1$ but not $v_2$. A belief update via the link containing $v_1$ will make sure that its belief is consistent in $CTF_k$ and $CTF_{k+1}$. Later, if we perform a belief update using a link containing $v_2$, the previous larger belief update of $v_1$ will be overwritten by something smaller since the belief of $v_2$ is not very different in the two CTFs. 

\textcolor{black}{
\textbf{Complexity}: 
Let $N_{CTF}$ be the number of CTFs in the sequence, and $N_l$ be the maximum number of selected links between any two adjacent CTFs. IBIA requires {$2 N_{CTF}$} rounds of message passing for calibrating CTFs and $(N_{CTF}-1)N_l$ rounds for belief update. Therefore, the runtime complexity of IBIA is $O(N_{CTF} N_l2^{mcs_p})$.
To perform backpropagation of beliefs, all CTFs need to be stored and the memory complexity is $O(N_{CTF} 2^{mcs_p})$.}

%% file: tikz/links.tex
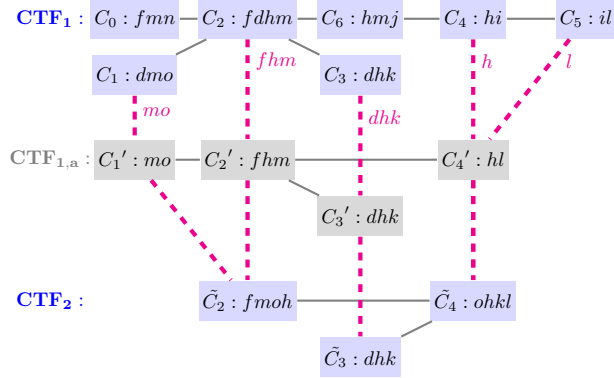
\begin{wrapfigure}{r}{0.58\linewidth}
\centering
    \begin{tikzpicture}[thick,scale=0.75, every node/.style={transform shape}]
    \node (ctf1) at (-2.5,0) {\color{blue}$\mathbf{CTF_1:}$};
        \node [selected vertexb](dhk3) at (3,-6) {$\tilde{C_3}:dhk$}; 
        \node [vertexr](dhk2) at (3,-3.5) {${C_3}':dhk$}; 
        \node [selected vertexb](dhk1) at (3,-1) {$C_3:dhk$}; 
        \foreach \pos/\name/\label in {(1,0)/{$C_2:fdhm$}/fdhm, (5,0)/{$C_4:hi$}/hi, (7,0)/{$C_5:il$}/il, (-1,-1)/{$C_1:dmo$}/dmo, (-1,0)/{$C_0:fmn$}/fmn, (3,0)/{$C_6: hmj$}/hmj}
    \node [selected vertexb] (\label) at \pos {\name};
    \path[edge2, gray] (fmn)--(fdhm)--(hmj)--(hi)--(il);
    \foreach \source/\dest in {fdhm/dmo, fdhm/dhk1}
    \path[edge2, gray] (\source)-- (\dest);
        \node (ctfa) at (-2.5,-2.5) {\color{gray}$\mathbf{CTF_{1,a}:}$};
        \foreach \pos/\name/\label in {(-1,-2.5)/{${C_1}':mo$}/mo, (1,-2.5)/{${C_2}':fhm$}/fhm, (5,-2.5)/{${C_4}':hl$}/hl}
    \node [vertexr] (\label) at \pos {\name};
    \foreach \source/\dest in {fhm/mo, fhm/dhk2, fhm/hl}
    \path[edge2,gray] (\source)-- (\dest);
    \node (ctfa) at (-2.5,-5) {\color{blue}$\mathbf{CTF_2:}$};
        \foreach \pos/\name/\label in {(1,-5)/{$\tilde{C_2}:fmoh$}/fmoh, (5,-5)/{$\tilde{C_4}:ohkl$}/ohkl}  
    \node [selected vertexb] (\label) at \pos {\name};
    \foreach \source/\dest in {fmoh/ohkl, ohkl/dhk3}
    \path[edge2,gray] (\source)-- (\dest);
    \foreach \source/\destx/\desty/\weight in { fhm/fdhm/fmoh/$fhm$, hl/hi/ohkl/$h$, dhk2/dhk1/dhk3/$dhk$}
    \path[udashededge] (\destx) -- node[right,midway, yshift=0.5cm]{\weight} (\source)-- (\desty);
    \path[udashededge] (dmo) -- node[right,midway, yshift=0.1cm]{$mo$} (mo)-- (fmoh);
    \path[udashededge] (il) -- node[right,midway, yshift=0.5cm, xshift=0.5cm]{$l$} (hl)-- (ohkl);
    \end{tikzpicture}
    \caption{Links between corresponding cliques in $CTF_1$, $CTF_{1,a}$ and $CTF_2$ for the example shown in Figure~\ref{fig:ibiaExample}.
        All links $(C,C',\tilde{C})$ are marked with dashed magenta lines and the link variables corresponding to each link are marked in magenta color. 
    }
    \label{fig:links}
\end{wrapfigure}

%% file: results.tex
\section{Results}
All experiments were carried out on an Intel i9-12900 Linux system running Ubuntu 22.04.

\textbf{Error metrics:} For each non-evidence variable $X_i$, we measure error in terms of \textit{Hellinger distance} between the exact marginal distribution $P(X_i)$ and the approximate marginal distribution $Q(X_i)$. It is computed as follows.
\begin{equation}
    HD = \frac{1}{\sqrt{2}}\sqrt{\sum\limits_{s\in Domain(X_i)}\{\sqrt{P(X_i=s)}-\sqrt{Q(X_i=s)}~\}^2} 
\end{equation}
We use two metrics namely, the \textit{average Hellinger distance} (denoted as $HD_{avg}$) and the \textit{maximum Hellinger distance} (denoted as $HD_{max}$) over all non-evidence variables in the network.

\textbf{Benchmarks: } We used the benchmark sets included in UAI repository~\citep{IhlerURL} and the Bayesian network repository~\citep{BnlearnURL}. 
We classify instances for which exact solutions are present in the repository as \textit{`small'} and others as \textit{`large'}.

\textbf{Methods used for comparison:}
\textcolor{black}{In the UAI 2022 inference competition~\citep{UAI2022}, the $uai14\_mar$ solver had the highest score for the MAR task. 
It is an amalgam of solvers that dumps solutions with different methods based on the given time and memory constraints. It uses loopy BP (LBP), generalized BP on loopy graphs where outer regions are selected using mini-bucket heuristics, and cutset conditioning of GBP approximations.
The implementation of this solver is not publicly available. Therefore, we have compared our results individually with methods that belong to categories of methods used in $uai14\_mar$. This includes LBP~\citep{Murphy1999}, IJGP~\citep{Mateescu2010} and sample search~\citep{Gogate2011} which is an importance sampling based technique that uses an IJGP based proposal and cutset sampling (referred to as `ISSwc' in this paper).
 We also compare our results with weighted mini-bucket elimination (WMB)~\citep{Liu2011}.
}
Additional results showing a comparison with results published in~\citet{Kelly2019} are included in the supplementary material.

\textbf{Evaluation setup:} 
\textcolor{black}{The implementation of LBP and WMB were taken from LibDAI~\citep{libdaiPaper,libdaiURL} and Merlin~\citep{Merlin} respectively.}
For IJGP and ISSwc, we used  implementations~\citep{ijgp,ijgpss} provided by the authors of these methods. LBP, IJGP, ISSwc and WMB are implemented in C++.
IBIA on the other hand has been implemented in Python3 and is thus, at a disadvantage in terms of runtime. 
We report results with runtime limits of 2 min and 20 min for small instances. In all cases, the memory limit was set to 8GB, which is the same as that used in UAI 2022 competition~\citep{UAI2022}. 
For IBIA, we set the maximum clique size bound $mcs_p$ to 20 (referred to as \textit{`IBIA20'}) when the time limit is 2 min and we set it to 23 (referred to as \textit{`IBIA23'}) when the time limit is 20 min. $mcs_{im}$ is empirically chosen as 5 less than $mcs_p$. 
The evaluation setup used for other methods is included in the supplementary material.

\input{tables/averageHD2}

\textbf{Results:} Table~\ref{tab:smallInst} has the results for the small benchmark sets. It reports the average of $HD_{avg}$ and $HD_{max}$ over all instances in each set. 
We compare results obtained using LBP, WMB, IJGP, IBIA20, IBIA23 and ISSwc for both time constraints.
 The minimum error obtained for each time limit is marked in bold.  
    IBIA20 and IBIA23 solve all small instances within 2 min and 20 min respectively. 
\textcolor{black}{In 2~min, the accuracy obtained with IBIA20 is better than all other solvers for most benchmarks. 
For ObjDetect, Segment and Protein, it is comparable to WMB, IJGP/ISSwc and LBP respectively, which give the least errors for these testcases. 
}
In 20 min, IBIA23 gives \textcolor{black}{lower or comparable errors} in all testcases except DBN and ObjDetect. For DBNs, ISSwc reduces to exact inference in most instances and hence error obtained is close to zero. For ObjDetect, IJGP gives the least error closely followed by IBIA23.
\textit{Note that for many benchmarks the accuracy obtained with IBIA20 in 2 min is either better than or comparable to the accuracy obtained with other solvers in 20 min.}

A comparison of IBIA with results published in ~\citet{Kelly2019} for Gibbs sampling with Rao-blackwellisation (ARB) and IJGP is included in the supplementary material. It is seen that error obtained with IBIA is lower than both methods in majority of the testcases.

For BN instances, Table~\ref{tab:mnVsBn} compares the results obtained using IBIA20 when CTFs are constructed by adding factors in the topological order (columns marked as `TP') with that obtained using a non-topological order (columns marked as `NTP'). 
We compare the maximum error in partition function (PR) and the average $HD_{max}$ over all instances in each benchmark set.
\input{tables/bnVsmn}
We observe that the topological ordering gives better accuracy for both PR and marginals in all testcases except Pedigree.
\textcolor{black}{The advantage of this ordering is that once all the evidence variables are added, no belief-update is needed is needed for the subsequent CTFs (using Theorem~\ref{thm:post}). So, the number of belief update steps is lower, resulting in lower errors. However, a drawback of this ordering is that it is rigid and it sometimes results in a larger number of CTFs in the sequence which could lead to larger errors if all the evidence variables are added in later CTFs.}
When no evidence variables are present (e.g. GridBN, Bnlearn), both runtime and memory complexity is lower with topological ordering since marginals are consistent in all CTFs (using Corollary~\ref{thm:prior}) and belief update is not needed.
The average runtime with and without topological ordering was 1s and 146s respectively for GridBN  and 0.3s and 1.3s for Bnlearn.

\input{tables/numInst.tex}
To evaluate the scalability of the proposed algorithm, we ran it for large networks where the exact solutions are not known. 
Table~\ref{tab:numInst.MAR} tabulates the percentage of large instances in each benchmark set that could be solved using IBIA within 2 min, 20 min and 60 min.
For this experiment, we start with $mcs_p$$=$$20$ and allow it to increase if incremental build results in a CTF with larger clique sizes. 
IBIA could solve all $large$ instances in benchmark sets BN, Promedas, ObjDetect and Segmentation and most instances in Protein within 20 min. For other benchmarks, additional instances could be solved when the runtime was increased to 60 min. 
The memory required for the remaining Grids, DBN, CSP and Protein instances is more than 8~GB. The increased memory usage is due to the following reasons. 
Firstly, all calibrated CTFs in the SLCTF need to be stored in order to allow for back-propagation of beliefs and the memory required increases with the number of CTFs. The average number of CTFs in the remaining Grid, DBN and CSP benchmarks is 22, 58 and 80 respectively.
Secondly, for benchmarks with large variable domain sizes, the number of variables present in each clique in a CTF is small. Therefore, approximation using exact and local marginalization becomes infeasible and the subsequent CTFs have clique sizes greater than $mcs_p$, which results in increased memory usage. This is seen in 9 out of 395 Protein instances and 12 out of 54 CSP instances.
In addition to memory, the remaining Type4b instances also require additional runtime.
This is because during belief update of each CTF, we perform one round of message passing for each selected link and the number of links is large in these instances. 

%% file: tables/averageHD2.tex
\begin{table*}[t]\centering
    \caption{Comparison of average $HD_{avg}$ and average $HD_{max}$ (shown in gray background) obtained using various inference methods with two runtime constraints, 2 min and 20 min. The minimum error obtained for each time limit is highlighted in bold. Entries are marked as ‘-’ where all instances could not be solved within the set time limit. The total number of instances solved by each method is shown in the last row. $ev_a$: average number of evidence variables, $v_a$: average number of variables, $f_a$: average number of factors, $w_a$: average induced width and $dm_a$: average of the maximum variable domain size.
    }
\label{tab:smallInst}
\small
    \setlength\tabcolsep{1.25pt}
    \begin{tabular}{llllllllllllll}\toprule
        \multirow{2}{*}{} &Total & \multicolumn{1}{c}{Average stats} &\multicolumn{5}{c}{2 min} &\multicolumn{5}{c}{20 min} \\\cmidrule(lr){3-3}\cmidrule(lr){4-8}\cmidrule(lr){9-13}
&\#Inst & $(ev_a,v_a,f_a,w_a,dm_a)$ &LBP &WMB&IJGP &ISSwc &IBIA20 &LBP &WMB&IJGP &ISSwc &IBIA23 \\\midrule
BN &97 &(76,637,637,28,10) &- &0.037 &- &- &\textbf{2E-4} &0.023 &0.025 &- &- &\textbf{6E-5} \\
\rowcolor{gray!20} & & &- &0.228 &- &- &\textbf{9E-3} &0.230 &0.170 &- &- &\textbf{2E-3} \\
GridBN &29 &(0,595,595,37,2) &0.075 &0.066 &0.011 &0.003 &\textbf{5E-6} &0.075 &0.048 &0.010 &0.001 &\textbf{2E-7} \\
\rowcolor{gray!20} & & &0.478 &0.416 &0.111 &0.051 &\textbf{7E-4} &0.478 &0.381 &0.094 &0.015 &\textbf{1E-4} \\
Bnlearn &26 &(0,256,256,7,16) &0.010 &0.005 &0.011 &0.012 &\textbf{7E-5} &0.010 &\textbf{5E-6} &0.008 &0.006 &7E-6 \\
\rowcolor{gray!20} & & &0.089 &0.021 &0.050 &0.064 &\textbf{0.002} &0.089 &\textbf{1E-4} &0.025 &0.028 &2E-4 \\
Pedigree &24 &(154,853,853,24,5) &0.075 &0.018 &0.035 &0.033 &\textbf{0.009} &0.075 &0.015 &0.033 &0.021 &\textbf{0.008} \\
\rowcolor{gray!20} & & &0.555 &0.253 &0.470 &0.292 &\textbf{0.204} &0.555 &\textbf{0.194} &0.446 &0.234 &0.198 \\
Promedas &64 &(7,618,618,21,2) &0.032 &0.055 &0.124 &0.030 &\textbf{0.013} &0.032 &0.043 &0.120 &0.021 &\textbf{0.010} \\
\rowcolor{gray!20} & & &0.168 &0.295 &0.504 &0.139 &\textbf{0.086} &0.168 &0.245 &0.487 &0.096 &\textbf{0.072} \\
DBN &36 &(653,719,14205,29,2) &- &0.069 &0.081 &\textbf{0.016} &0.020 &- &0.018 &0.060 &\textbf{2E-6} &0.003 \\
\rowcolor{gray!20} & & &- &0.883 &0.919 &0.766 &\textbf{0.261} &- &0.544 &0.879 &\textbf{2E-4} &0.098 \\
ObjDetect &79 &(0,60,210,6,16) &0.022 &\textbf{0.001} &0.004 &0.018 &0.002 &0.022 &2E-4 &\textbf{3E-5} &0.009 &4E-4 \\
\rowcolor{gray!20} & & &0.130 &\textbf{0.010} &0.037 &0.189 &0.020 &0.130 &0.003 &\textbf{3E-4} &0.061 &0.006 \\
Grids &8 &(0,250,728,22,2) &0.433 &0.146 &0.247 &- &\textbf{0.088} &0.433 &0.089 &0.123 &0.056 &\textbf{0.002} \\
\rowcolor{gray!20} & & &0.905 &0.343 &0.713 &- &\textbf{0.300} &0.905 &0.221 &0.423 &0.209 &\textbf{0.099} \\
CSP &12 &(0,73,369,12,4) &0.019 &0.033 &0.026 &- &\textbf{0.002} &0.019 &0.022 &0.017 &0.054 &\textbf{2E-4} \\
\rowcolor{gray!20} & & &0.066 &0.101 &0.134 &- &\textbf{0.011} &0.066 &0.057 &0.073 &0.093 &\textbf{0.003} \\
Segment &50 &(0,229,851,17,2) &0.035 &1E-4 &\textbf{5E-6} &\textbf{5E-6} &6E-5 &0.035 &5E-6 &5E-6 &5E-6 &\textbf{1E-7} \\
\rowcolor{gray!20} & & &0.258 &0.002 &\textbf{7E-5} &\textbf{7E-5} &0.001 &0.258 &7E-5 &7E-5 &7E-5 &\textbf{4E-7} \\
Protein &68 &(0,59,176,6,77) &\textbf{5E-4} &0.005 &0.003 &0.003 &6E-4 &5E-4 &0.003 &0.007 &0.001 &\textbf{3E-5} \\
\rowcolor{gray!20} & & &\textbf{0.007} &0.102 &0.094 &0.049 &0.039 &0.007 &0.066 &0.230 &0.015 &\textbf{0.002} \\
        \midrule 
\#Inst &493 & &481 &493 &487 &485 &493 &485 &493 &491 &487 &493 \\
\bottomrule
\end{tabular}
\end{table*}

%% file: tables/bnVsmn.tex
\begin{wraptable}{r}{0.6\linewidth}
    \centering
    \caption{Comparison of maximum error in PR and average $HD_{max}$ obtained using IBIA20 with CTFs constructed by adding factors in topological order (shown in columns marked `TP') and that obtained using a non-topological order (shown in columns marked `NTP'). 
    $ev_a$: Average number of evidence variables, $\Delta_{PR}=~|~\log_{10} PR - \log_{10} PR^{*}~|$ where $PR$ and $PR^{*}$ are estimated and exact values.
}\label{tab:mnVsBn}
\small
\begin{tabular}{lrrrrrrr}\toprule
    & & &\multicolumn{2}{c}{Max $\Delta_{PR}$} &\multicolumn{2}{c}{Avg $HD_{max}$} \\\cmidrule(lr){4-5} \cmidrule(lr){6-7}
&\#Inst &$ev_a$ &NTP &TP &NTP &TP \\\midrule
Bnlearn &26 &0 &0.02 &\textbf{0} &0.023 &\textbf{0.002} \\
GridBN &29 &0 &0.09 &\textbf{0} &0.231 &\textbf{0.001} \\
Promedas &64 &7 &1.5 &\textbf{0.4} &0.322 &\textbf{0.086} \\
BN &97 &76 &0.07 &\textbf{0.02} &0.116 &\textbf{0.009} \\
Pedigree &24 &159 &\textbf{0.4} &0.7 &\textbf{0.098} &0.204 \\
\bottomrule
\end{tabular}
\end{wraptable}

%% file: tables/numInst.tex
\begin{wraptable}{r}{0.65\linewidth}
\vspace*{-0.25cm}
\centering
\caption{Percentage of large instances in each benchmark set solved by IBIA within 2, 20 and 60 minutes. $ev_a$: average number of evidence variables, $v_a$: average number of variables, $f_a$: average number of factors, $w_a$: average induced width and $dm_a$: average of the maximum domain-size.}\label{tab:numInst.MAR}
\small
\setlength\tabcolsep{2pt}
\begin{tabular}{lrrrrrrr}\toprule
    &Total & \multicolumn{1}{c}{Average stats} & \multicolumn{3}{c}{Instances solved (\%)} \\\cmidrule(lr){3-3} \cmidrule(lr){4-6}
&\#Inst &$(ev_a,v_a,f_a,w_a,dm_a)$ &2 min &20 min &60 min \\\midrule
BN &22 &(188,1272,1272,51,17) &64 &100 &100 \\
Promedas &171 &(15,1207,1207,71,2) &77 &100 &100 \\
ObjDetect &37 &(0,60,1830,59,17) &27 &100 &100 \\
Segment &50 &(0,229,851,19,21) &100 &100 &100 \\
Protein &395 &(0,306,1192,21,81) &75 &97 &98 \\
DBN &78 &(784,944,47206,60,2) &38 &77 &77 \\
Grids &19 &(0,3432,10244,117,2) &16 &37 &58 \\
CSP &54 &(0,294,11725,175,41) &31 &54 &59 \\
Type4b &82 &(4272,10822,10822,24,5) &0 &9 &29 \\
\bottomrule
\end{tabular}
\end{wraptable}

%% file: conclusions.tex
\section{Discussion}

\textbf{Limitations:} 
While the belief update algorithm performs well for most benchmarks, it has some limitations. It is sequential and is performed link by link for each CTF that needs to be updated. The time and space complexity depends on the number of CTFs in the sequence and the number of selected links, which is large in some testcases. 
Also, after belief-update of all CTFs is completed, beliefs of variables present in multiple CTFs need not be consistent. 
However, good accuracies are obtained when beliefs are inferred from the last CTF containing the variable.
For BNs, we found that building CTFs in the topological order gives larger errors in some cases. A possible extension would be to have an efficient build strategy where the ordering is decided dynamically based on the properties of the graph structure.

\textbf{Comparison with related work}: 
IBIA is similar to mini-bucket schemes in the sense that the accuracy-complexity tradeoff is controlled using a user-defined maximum clique size bound. While mini-bucket based schemes like IJGP~\citep{Dechter02} and join graph linear programming~\citep{Ihler2012} use iterative message passing in loopy graphs, others like mini-bucket elimination (MBE), WMB~\citep{Liu2011} and mini-clustering~\citep{Mateescu2010} are non-iterative approaches that approximate the messages by migrating the sum operator inside the product term. In contrast, IBIA constructs a sequence of clique trees. It performs belief propagation on approximate clique trees so that messages are exact and there are no issues of convergence.

Unlike sampling based techniques, there is no inherent randomness in IBIA that is each run gives the same results. 
There could be a variation if the order in which factors are added is changed. However, this variation is minimal since 
subsets of factors are added together in the incremental build step.
In that sense, it is like mini-bucket based methods where results are the same if the variable elimination order and the partitioning technique used to generate the initial choice of mini-buckets are the same.

In order to construct the sequence, IBIA requires a fast and accurate method to approximate clique trees by reducing clique sizes. The aim is to preserve as much as possible, the joint distribution of the interface variables.
This is achieved by marginalizing out variables from large-sized cliques and a minimal set of neighbors without disconnecting clique trees. IBIA gives good accuracy since variables that are removed are chosen based on a mutual information (MI) based metric and a sufficient number of non-interface variables are retained so that a CT is never disconnected. In contrast, the Boyen Koller (BK)~\citep{Boyen1998,Murphy2002} and factored frontier (FF) approximations, although fast, retain only the interface variables which can disconnect the CT resulting in larger errors due to the underlying independence approximation. The thin junction tree approximations proposed in \citet{Kjaerulff1994} and \citet{Hutter2004} split cliques that contain a pair of variables that is not present in any of its sepsets. However, if there are large cliques that have only sepset variables (which is typical in most benchmarks), then the split has to be done iteratively starting from leaf nodes of multiple branches, until the large-sized clique can be split. When such cliques are centrally located in the CT, this process is both time-consuming and would result in approximation of a much larger set of cliques. Similar to BK and FF, this method can also disconnect the CT.

The other thin junction tree methods~\citep{Bach2001,Elidan2008,Dafna2009} choose an optimum set of features based on either KL distance or a MI based metric. They cannot be directly used since IBIA requires a good approximation of the joint distribution of only the interface variables. Also, these methods are typically iterative and not very fast.

%% file: supplementary.tex


\appendix
\section{Results}
\subsection{Evaluation setup}
For loopy belief propagation (LBP)~\citep{Murphy1999}, we use the implementation provided in LibDAI~\citep{libdaiPaper,libdaiURL}. We set the tolerance limit to $10^{-3}$ when time limit is 2 min and $10^{-9}$ for 20 min. For iterative join graph propagation (IJGP)~\citep{Mateescu2010}, we used the implementation available on the author's webpage~\citep{ijgp}. The maximum cluster size in IJGP is set using the parameter $ibound$. This solver starts with the minimal value of $ibound$ and increases it until the runtime and memory constraints are satisfied. A solution is obtained for each $ibound$. \textcolor{black}{The results reported are those obtained for the largest $ibound$ possible for the given time and memory constraints.}
For WMB, we used the implementation made available by the authors in the Merlin tool~\citep{Merlin}. Since this implementation uses a fixed $ibound$ value, we \textcolor{black}{wrote a script to run} it in anytime fashion similar to IJGP. We report results obtained with the largest value of $ibound$ possible.
For sample search with IJGP-based proposal and cutset sampling (ISSwc)~\citep{Gogate2011}, we used the implementation provided by the authors on Github~\citep{ijgpss}.  For ISSwc, appropriate values of $ibound$ and $w\mbox{-}cutset$ bound are set by the tool based on the given runtime limit.
\subsection{Additional results}
For a fair comparison with IBIA using $mcs_p$ of $20$ (referred to as `IBIA20'), we also obtained the results for ISSwc after fixing both $ibound$ and $w\mbox{-}cutset$ bound to 20 (referred to as `ISSwc20').
\textcolor{black}{Table~\ref{tab:ijgpss} compares the results obtained using IBIA20, ISSwc20 and ISSwc (in which the optimal $ibound$ is determined by the solver).}
The runtime limit was set to 2 min and 20 min, and the memory limit was set to 8~GB.
The error obtained using IBIA20 is either smaller than or comparable to ISSwc20 and ISSwc for both time limits in all testcases except DBN. 
For DBN, in 2 min, the average $HD_{max}$ obtained with IBIA20 is significantly smaller than both variants of sample search, and the average $HD_{avg}$ obtained with IBIA20 is comparable.
However, in 20 min, both variants reduce to exact inference in many DBN instances and the average error obtained is close to zero.

Table~\ref{tab:arb} compares the maximum Hellinger distance obtained using IBIA ($mcs_p$=15,20) with published results for adaptive Rao Blackwellisation (ARB) and iterative join graph propagation in \citet{Kelly2019}. The minimum error obtained is shown in bold. IBIA with $mcs_p=20$ gives the least error in all cases. The error obtained with $mcs_p=15$ is smaller than ARB and IJGP in all testcases except Grids\_11, Grids\_13 and Promedas\_12.
\input{tables/ijgpss1}
\input{tables/arb}
\newpage

\section{Pseudo-code} 
Algorithm~\ref{alg:inferMAR} shows the steps in the proposed algorithm for the inference of marginals. We first convert the PGM into a sequence of linked CTFs ($SLCTF$) that contains a sequence of calibrated CTFs ($SCTF=\{CTF_k\}$) and a list of links between adjacent CTFs ($SL=\{L_k\}$). Functions $BuildCTF$ and $ApproximateCTF$ are used for incremental construction of CTFs and approximation of CTFs respectively. The steps in these functions are explained in detail in Algorithms 1 and 2 in~\citet{IBIAPR}. Links between adjacent CTFs are found using the function $FindLinks$ and belief update in the SLCTF is performed using the function $BeliefUpdate$. Following this, the marginal of a variable $v$ is inferred from clique beliefs in the last CTF that contains $v$ (line 23).
\input{Algo_inferMAR}


\input{proofs1}

%% file: tables/ijgpss1.tex
\begin{table}[!htp]\centering
    \caption{Comparison of average $HD_{avg}$ and average $HD_{max}$ (shown in gray background) obtained using IBIA with $mcs_p=20$ (IBIA20), ISSwc with clique size bounds determined by the solver~\citep{ijgpss} (ISSwc) and ISSwc with $ibound$ and $w\mbox{-}cutset$ bound fixed to 20 (ISSwc20). Results are shown for two runtime limits, 2 min and 20 min.  Entries are marked with `-' if the solution for all testcases could not be obtained within the given time and memory limits. The minimum error obtained for a benchmark is highlighted in bold.
    The number of instances solved by each solver is shown in the last row. $ev_a$: average number of evidence variables, $v_a$: average number of variables, $f_a$: average number of factors, $w_a$: average induced width and $dm_a$: average of the maximum variable domain size.
    }\label{tab:ijgpss}
\small
\setlength\tabcolsep{3pt}
\begin{tabular}{lrrrrrrrrr}\toprule
\multirow{2}{*}{} &Total &\multirow{2}{*}{$(ev_a,v_a,f_a,w_a,dm_a)$} &\multicolumn{3}{c}{2 min} &\multicolumn{2}{c}{20 min} & \\ \cmidrule(lr){4-6} \cmidrule(lr){7-9}
&\#Inst & &ISSwc &ISSwc20 &IBIA20 &ISSwc &ISSwc20 &IBIA20 \\\midrule
BN &97 &(76,637,637,28,10) &- &0.037 &\textbf{0} &- &0.033 &\textbf{0} \\
\rowcolor{gray!20} & & &- &0.145 &\textbf{0} &- &0.085 &\textbf{0} \\
GridBN &29 &(0,595,595,37,2) &0.003 &0.005 &\textbf{0} &0.001 &0.005 &\textbf{0} \\
\rowcolor{gray!20} & & &0.051 &0.065 &\textbf{0} &0.015 &0.046 &\textbf{0} \\
Bnlearn &26 &(0,256,256,7,16) &0.012 &0.036 &\textbf{0} &0.006 &0.036 &\textbf{0} \\
\rowcolor{gray!20} & & &0.064 &0.094 &\textbf{0.002} &0.028 &0.093 &\textbf{0.002} \\
Pedigree &24 &(154,853,853,24,5) &0.033 &0.028 &\textbf{0.009} &0.021 &0.021 &\textbf{0.009} \\
\rowcolor{gray!20} & & &0.292 &0.245 &\textbf{0.204} &0.234 &\textbf{0.195} &0.204 \\
Promedas &64 &(7,618,618,21,2) &0.030 &0.042 &\textbf{0.013} &0.021 &0.033 &\textbf{0.013} \\
\rowcolor{gray!20} & & &0.139 &0.207 &\textbf{0.086} &0.096 &0.153 &\textbf{0.086} \\
DBN &36 &(653,719,14205,29,2) &0.016 &\textbf{0.011} &0.020 &\textbf{0} &\textbf{0} &0.020 \\
\rowcolor{gray!20} & & &0.766 &0.833 &\textbf{0.261} &\textbf{0} &\textbf{0} &0.261 \\
ObjDetect &79 &(0,60,210,6,16) &0.018 &0.039 &\textbf{0.002} &0.009 &0.004 &\textbf{0.002} \\
\rowcolor{gray!20} & & &0.189 &0.233 &\textbf{0.020} &0.061 &0.021 &\textbf{0.020} \\
Grids &8 &(0,250,728,22,2) &- &- &\textbf{0.088} &\textbf{0.056} &- &0.088 \\
\rowcolor{gray!20} & & &- &- &\textbf{0.300} &\textbf{0.209} &- &0.300 \\
CSP &12 &(0,73,369,12,4) &- &- &\textbf{0.002} &0.054 &0.069 &\textbf{0.002} \\
\rowcolor{gray!20} & & &- &- &\textbf{0.011} &0.093 &0.081 &\textbf{0.011} \\
Segment &50 &(0,229,851,17,2) &\textbf{0} &0.002 &\textbf{0} &\textbf{0} &\textbf{0} &\textbf{0} \\
\rowcolor{gray!20} & & &\textbf{0} &0.036 &0.001 &\textbf{0} &\textbf{0} &0.001 \\
Protein &68 &(0,59,176,6,77) &0.003 &0.003 &\textbf{0} &0.001 &0.001 &\textbf{0} \\
\rowcolor{gray!20} & & &0.049 &\textbf{0.030} &0.039 &0.015 &\textbf{0.011} &0.039 \\
\midrule 
\#Inst &493 & &485 &488 &493 &487 &489 &493 \\
\bottomrule
\end{tabular}
\end{table}

%% file: tables/arb.tex
\begin{table}[!htp]\centering
    \caption{Comparison of maximum Hellinger distance ($HD_{max}$) obtained using IBIA with published results for Gibbs sampling with adaptive Rao Blackwellisation (ARB) and iterative join graph propagation in \citet{Kelly2019}. Results obtained with $mcs_p=15$ and $mcs_p=20$ are shown in columns marked as IBIA15 and IBIA20 respectively. Runtimes (in seconds) for IBIA15 and IBIA20 are also shown. Estimates for ARB were obtained within 600 seconds$^{+}$~\citep{Kelly2019} and runtime for IJGP is not reported in~\citet{Kelly2019}. The minimum error obtained for each benchmark is marked in bold. $w$: induced width, $dm$: maximum domain size }\label{tab:arb}
\small
\begin{tabular}{lccccccccc}\toprule
    & & &\multicolumn{4}{c}{$HD_{max}$} &\multicolumn{2}{c}{Runtime (s)} \\\cmidrule(lr){4-7} \cmidrule(lr){8-9}
    &$w$ &$dm$ &Merlin (IJGP)$^{*}$ &ARB$^{*}$ &IBIA15 &IBIA20 &IBIA15 &IBIA20 \\\midrule
Alchemy\_11 &19 &2 &0.777 &0.062 &0.004 &\textbf{1E-7} &3.3 &2.9 \\
CSP\_11 &16 &4 &0.513 &0.274 &0.100 &\textbf{0.034} &0.5 &3.4 \\
CSP\_12 &11 &4 &0.515 &0.275 &0.028 &\textbf{6E-7} &0.1 &0.1 \\
CSP\_13 &19 &4 &0.503 &0.290 &0.085 &\textbf{0.051} &0.9 &2.9 \\
Grids\_11 &21 &2 &0.543 &0.420 &0.590 &\textbf{0.166} &1.1 &3.5 \\
Grids\_12 &12 &2 &0.645 &0.432 &\textbf{3E-7} &\textbf{3E-7} &0.0 &0.0 \\
Grids\_13 &21 &2 &0.500 &0.544 &0.962 &\textbf{0.246} &1.1 &3.6 \\
Pedigree\_11 &19 &3 &0.532 &0.576 &0.016 &\textbf{5E-7} &0.5 &0.1 \\
Pedigree\_12 &19 &3 &0.562 &0.506 &0.023 &\textbf{4E-7} &0.3 &0.1 \\
Pedigree\_13 &19 &3 &0.577 &0.611 &5E-7 &\textbf{5E-7} &0.1 &0.1 \\
Promedus\_11 &18 &2 &1.000 &0.373 &0.049 &\textbf{5E-7} &1.4 &0.5 \\
Promedus\_12 &20 &2 &1.000 &0.358 &0.657 &\textbf{0.242} &2.8 &4.1 \\
Promedus\_13 &10 &2 &1.000 &0.432 &\textbf{5E-7} &\textbf{5E-7} &0.4 &0.4 \\
\bottomrule
\end{tabular}
    {\\ \scriptsize $^{*}$ The results tabulated in~\citet{Kelly2019} report -$\log_2 HD_{max}$. The table above has the corresponding values of $HD_{max}$.\\
    $^{+}$ System used: Ubuntu 18.04, with 16GB of RAM, 6 CPUs and 2 hardware threads per CPU~\citep{Kelly2019}.
    }
\end{table}

%% file: Algo_inferMAR.tex
\begin{algorithm}[!ht]
    \small
    \caption{InferMarginals ($\Phi, mcs_p, mcs_{im}$)}
	\label{alg:inferMAR}
	\begin{algorithmic}[1]
            \Require $\Phi$: Set of factors in the PGM \newline
            \indent$mcs_p$: Maximum clique size bound for each CTF in the sequence \newline
            \indent$mcs_{im}$: Maximum clique size bound for the approximate CTF            
            \Ensure $MAR$: Map containing marginals $<~variable: margProb~>$
		\State Initialize: $MAR = <>$ \Comment{{\color{teal!70}  Map $<variable: margProb>$}}\newline
                           \indent\indent $S_v=\cup_{\phi\in\Phi} Scope(\phi)$ \Comment{{\color{teal!70}Set of all variables in the PGM}}\newline
                           \indent\indent $SCTF=[~]$ \Comment{{\color{teal!70} Sequence of calibrated CTFs}}\newline
                           \indent\indent $SL=[~]$ \Comment{{\color{teal!70} List of list of links between all adjacent CTFs}}\newline
                            \indent\indent$k=1$ \Comment{{\color{teal!70}Index of CTF in $SCTF$}}
            \While{$\Phi.isNotEmpty()$}
            \Comment{{\color{teal!70} Convert PGM $\Phi$ to $SLCTF=\{SCTF, SL\}$}}
            \If{$k==1$}
                \State $CTF_0\gets$ Disjoint cliques corresponding to factors in $\Phi$ with disjoint scopes
                \LineComment{{\color{teal!70}Add factors to $CTF_0$ using BuildCTF (Algorithm~1 in~\citet{IBIAPR})}}
                \State $CTF_1, \Phi_1 \gets $ BuildCTF ($CTF_0, \Phi, mcs_p$) \Comment{{\color{teal!70} $\Phi_1$: Subset of factors in $\Phi$  added to $CTF_1$}} 
                \State $\Phi\gets\Phi\setminus\Phi_1$ \Comment{{\color{teal!70}Remove factors added to $CTF_1$ from $\Phi$}}
            \Else
                \LineComment{{\color{teal!70}Add factors to $CTF_{k-1,a}$ using BuildCTF (Algorithm~1  in~\citet{IBIAPR})}}
                \State $CTF_k, \Phi_k \gets $ BuildCTF ($CTF_{k-1,a}, \Phi, mcs_p$) \Comment{{\color{teal!70} $\Phi_k$: Subset of factors in $\Phi$  added to $CTF_k$}} 
                \State $\Phi\gets\Phi\setminus\Phi_k$ \Comment{{\color{teal!70}Remove factors added to $CTF_k$ from $\Phi$}}
                \State $L_{k-1} \gets$ FindLinks($CTF_{k-1}, CTF_{k-1,a},CTF_k$) \Comment{{\color{teal!70}$L_{k-1}$: List of links between $CTF_{k-1}$,$CTF_k$}}
                \State $SL.append(L_{k-1})$ \Comment{{\color{teal!70} Add $L_{k-1}$ to the sequence of links $SL$}}
            \EndIf
            \State Calibrate $CTF_k$ using belief propagation
            \State $SCTF.append(CTF_k)$ \Comment{{\color{teal!70}Add $CTF_k$ to the sequence $SCTF$}}
            \LineComment{{\color{teal!70}Reduce clique sizes to $mcs_{im}$ using ApproximateCTF (Algorithm 2 in~\citet{IBIAPR})}}
            \State $CTF_{k,a}\gets$ ApproximateCTF ($CTF_k,\Phi,mcs_{im}$) 
            \State $k\gets k+1$
            \EndWhile
            \State $SLCTF= \{SCTF,SL\}$ \Comment{{\color{teal!70} Sequence of linked CTFs}}
            \State BeliefUpdate($SLCTF$) \Comment{{\color{teal!70} Re-calibrate CTFs so that beliefs in all CTFs account for all factors}}
            \State $MAR[v]\gets$ Find marginal of $v$ from $CTF_j$ s.t. $v\in CTF_{k}, v \not\in CTF_{k+1}~~~~~\forall v\in S_v$   \Comment{{\color{teal!70} Infer marginals}}
        \State
        \Procedure{FindLinks}{$CTF_{k-1}, CTF_{k-1,a}, CTF_{k}$}
                    \LineComment{{\color{teal!70} Each link is a triplet consisting of $C\in CTF_{k-1}$, $C'\in CTF_{k-1,a}$ and $\tilde{C}\in CTF_k$}}
                    \For {$C'\in CTF_{k-1,a}$} \Comment{{\color{teal!70}Find links corresponding to each clique $C'$ in $CTF_{k-1,a}$}}
                    \LineComment{{\color{teal!70}Find list of corresponding cliques in $CTF_{k-1}$, $L_c$ }}
                    \If {$C'.isCollapsedClique$} \Comment{{\color{teal!70}$C'$ is obtained after exact marginalization}}
                    \State $L_c\gets$ List of cliques in $CTF_{k-1}$ that were collapsed to form $C'$
                    \Else \Comment{{\color{teal!70}$C'$ is either obtained after local marginalization or it is present as is in $CTF_k$}}
                    \State $C\gets $ Clique in $CTF_{k-1}$ s.t. $C'\subseteq C$; $L_c=[C]$
                    \EndIf
                    \State Find clique $\tilde{C}$ in $CTF_{k}$ s.t. $C'\subseteq\tilde{C}$
                    \LineComment{{\color{teal!70}Add all links corresponding to $C'$}}
                    \State \textbf{for} {$C\in L_c$} \textbf{do}        $L_{k-1}.append((C, C', \tilde{C}))$ \textbf{end for}                    
                \EndFor
            \State \Return $L_{k-1}$
        \EndProcedure
        \State
        \Procedure{BeliefUpdate}{$SLCTF$}
            \State $SCTF,SL=SLCTF$
            \For {$k \in len(SCTF)$~ down~ to~ 2}
            \Comment{{\color{teal!70}  Update beliefs in $\{CTF_k, k<len(SCTF)\}$}}
                \State $CTF_{k-1} \gets SCTF[k-1]$; $CTF_{k} = SCTF[k]$; $L_{k-1}=SL[k-1]$
                \State $L_{s} \gets $Priority queue with subset of links in $L_{k-1}$ chosen using heuristics described in Section 3.2
                \For {$(C, C', \tilde{C}) \in L_s$}
        \Comment{{ \color{teal!70}Back-propagate beliefs from $CTF_k$ to $CTF_{k-1}$ via all selected links}}
                    \State $\beta(C) =  \frac{\beta(C)}{\sum\limits_{C\setminus \{C\cap C'\}}\beta(C)} ~  \sum \limits_{{\tilde{C}} \setminus \{C\cap C'\}} \beta(\tilde{C})$ \Comment{{\color{teal!70} Update $\beta(C)\in CTF_{k-1}$ based on $\beta(\tilde{C})  \in CTF_{k}$}}
                    \State Update belief of all other cliques in $CTF_{k-1}$ using single pass message passing  with $C$ as root 
                \EndFor
            \EndFor
        \EndProcedure
    \end{algorithmic}
\end{algorithm}

%% file: proofs1.tex
\section{Proofs}

\textbf{Notations}
\newlist{indenteddesc}{description}{1}
\setlist[indenteddesc]{
  leftmargin=4.5em,  
  rightmargin=0em,
  labelindent=0em, 
  labelwidth=4em,
  labelsep=.5em
}
\begin{indenteddesc}
    \itemsep0em
    \item [$\Phi_k$] Set of factors added \textcolor{black}{to construct} $CTF_k$
    \item [$X_k$] Set of all \textcolor{black}{non-evidence} variables in $CTF_k$
    \item [$X_{k,a}$] Set of all \textcolor{black}{non-evidence} variables in $CTF_{k,a}$
    \item [$Y_k$] \textcolor{black}{Set of variables in $CTF_k$ but not in $CTF_1,\hdots,CTF_{k-1}$}
    \item [$Pa_{Y_k}$] Parents of variables in $Y_k$ in the BN
    \item [$E_k$] \textcolor{black}{Set of evidence variables in $Y_k$}
\item [$e_k$] \textcolor{black}{Evidence state corresponding to  variables in $E_k$} 
      \item [$C$] A clique in $CTF_k$
  
    \item [$C'$] A clique in $CTF_{k,a}$
        \item [$SP$] Sepset associated with an edge in $CTF_k$
    \item [$SP'$] Sepset associated with an edge in $CTF_{k,a}$
     {\color{black}
   \item [$\beta(C)$] Unnormalized clique belief of clique $C$
   \item [$\beta_N(C)$] Normalized clique belief of clique $C$, $\beta_N(C)=\frac{\beta(C)}{\sum\limits_{v\in C}\beta(C)}$ 
   }

    \item [$Z_k$] Normalization constant of the distribution encoded by calibrated beliefs in $CTF_k$
    \item [$\textcolor{black}{Q_k}(X_k)$] Probability distribution corresponding to $CTF_k$
    \item [$\textcolor{black}{Q_{k,a}}(X_{k,a})$] Probability distribution corresponding to $CTF_{k,a}$
\end{indenteddesc}

\underline{\textbf{Propositions related to inference of marginals:}} Let $CTF_k$ be a CTF in the SCTF generated by the IBIA framework and $CTF_{k,a}$ be the corresponding approximate CTF. 

\begin{proposition}\label{prop:appBeliefs}
    The joint belief of variables contained within any clique in the approximate CTF $CTF_{k,a}$ is the same as that in $CTF_k$.
\end{proposition}
\begin{proof}
\color{black}
    The approximation algorithm has two steps, exact marginalization and local marginalization.
    Exact marginalization involves finding the joint belief by collapsing all cliques containing a variable and then marginalizing the belief by summing over the states of the variable. This does not change the belief of the remaining variables. Local marginalization involves marginalizing a variable from individual cliques and sepsets by summing over its states. Let $C'$ denote the clique obtained after local marginalization of variable $v$ from clique $C$. 
    The updated clique belief ($\beta(C')$) is computed as shown below.
        \begin{align*}
            \beta(C') = \sum\limits_{v}\beta(C) 
        \end{align*}
        Once again, summing over the states of a variable does not alter the joint belief of the remaining variables in the clique.

\end{proof}


\newpage
\begin{proposition}\label{prop:ctfkBeliefs}
    The clique beliefs in $CTF_k$ account for all factors added to \{$CTF_1, \hdots,CTF_k$\}.
\end{proposition}
\begin{proof} 
\color{black}
$CTF_1$ is constructed by adding factors to an initial CTF that contains a set of disjoint cliques corresponding to a subset of factors with disjoint scopes. Let $\Phi_1$ be the set of all factors present in $CTF_1$ and $Z_1$ be the corresponding normalization constant. 
After calibration, the normalized clique belief ($\beta_N(C)$) of any clique $C$ in $CTF_1$ can be computed as follows.
    \begin{align*}
        \beta_N(C) = \frac{1}{Z_1} \sum\limits_{X_1\setminus C} \frac{\prod_{C_i\in CTF_1}\beta(C_i)}{\prod_{SP\in CTF_1}\mu(SP)} &= \frac{1}{Z_1} \sum\limits_{X_1\setminus C} \prod\limits_{\phi\in \Phi_1} \phi
    \end{align*}
 Therefore, clique beliefs in $CTF_1$ account for all factors in $\Phi_1$. 
 
$CTF_{1,a}$ is a calibrated CTF (refer Proposition~6,~\citet{IBIAPR}) that is obtained after 
\textcolor{black}{approximate marginalization of the variables in $X_1\setminus X_{1,a}$. Therefore, the joint distribution of variables in $CTF_{1,a}$ also accounts for all factors in $\Phi_1$.}
$CTF_{2}$ is constructed by adding factors in $\Phi_2$ to $CTF_{1,a}$. 
Therefore, after calibration, the normalized clique belief ($\beta_N(C)$) of any clique $C$ in $CTF_2$ can be computed as follows.
    \begin{align}\label{eq:betac21}
        \beta_N(C) &= \frac{1}{Z_2} \sum\limits_{X_2\setminus C} \frac{\prod_{C'\in CTF_{1,a}}\beta(C')}{\prod_{SP'\in CTF_{1,a}}\mu(SP')} \prod\limits_{\phi\in \Phi_2} \phi
    \end{align}
    where, $Z_2$ is the normalization constant of the distribution in $CTF_2$.
    \textcolor{black}{Using equation \ref{eq:betac21},  the clique beliefs in $CTF_{2}$ accounts for all factors in $\Phi_1$ and $\Phi_2$.}
    

A similar procedure can be repeated for subsequent CTFs to show that the proposition holds true for all CTFs in the sequence.
\end{proof}

\underline{\textbf{Propositions related inference in BNs:}}
\color{black}

The following propositions hold true for Bayesian networks when \textbf{each CTF in the SCTF is constructed by adding factors or conditional probability distributions (CPD) of variables in the topological order}.
$Y_k$ denotes the set of variables whose CPDs are added during construction of $CTF_k$ and $e_k$ denotes the evidence states of all evidence variables in $Y_k$.
\color{black}
\begin{proposition}\label{prop:tp}
The product of factors added in CTFs, $\{CTF_1,\hdots,CTF_k\}$ is a valid \textcolor{black}{joint} probability distribution whose normalization constant is the probability of evidence states $e_1,\hdots,e_k$. 
\end{proposition}

\begin{proof}
\textcolor{black}{Let $\mathcal{Y}_k=\{Y_1,\hdots,Y_k\}$ and $\varepsilon_k=\{e_1,\hdots,e_k\}$. Since CTFs are constructed by adding CPDs of variables in the topological order, the CPDs of parents $Pa_{Y_k}$ are present in $\{CTF_1,\hdots,CTF_k\}$. Therefore, the product of the CPDs is the unnormalized joint probability distribution $P(\mathcal{Y}_k,\varepsilon_k)$. Since the CPDs of all non-evidence variables are normalized to one, the normalization constant is $P(\varepsilon_k)$.}
\end{proof}

%

\begin{proposition}\label{prop:NC}
The normalization constant of the distribution encoded by the calibrated beliefs in $CTF_k$ is the estimate of probability of evidence states $e_1, \hdots, e_k$.
\end{proposition}
\begin{proof}
    The initial factors assigned to $CTF_{1}$ are CPDs of variables in $Y_1$. 
    Therefore, using Proposition~\ref{prop:tp}, the NC obtained after calibration is $Z_1=P(e_1)$.

 $CTF_{1,a}$ is obtained after approximation of $CTF_1$. 
 \textcolor{black}{All CTs in $CTF_{1,a}$ are calibrated CTs and the normalization constant of the distribution in $CTF_{1,a}$ is same as that of $CTF_1$ (refer Propositions~6 and 9 in \citet{IBIAPR}.}
    However, due to local marginalization, the overall distribution represented by $CTF_{1,a}$ is approximate.
The probability distribution corresponding to $CTF_{1,a}$ can be written as follows.
 \begin{align}\label{eq:appD}
\textcolor{black}{Q_{1,a}}(X_{1,a}|e_1) &= \frac{1}{Z_1} \frac{\prod_{C'\in CTF_{1,a}}\beta(C')}{\prod_{SP'\in CTF_{k,a}}\mu(SP')} \nonumber \\
 \implies Z_1 \textcolor{black}{Q_{1,a}}(X_{1,a}|e_1)       &= \textcolor{black}{Q_{1,a}}(X_{1,a},e_1)
 \end{align}
    where $X_{1,a}$ is the set of variables in $CTF_{1,a}$.

    $CTF_2$ is obtained after adding a new set of CPDs of variables in $Y_2$ to $CTF_{1,a}$. Let  \textcolor{black}{$X_2=X_{1,a}\cup\{Y_2\setminus E_2\}$} denote the set of  \textcolor{black}{non-evidence variables} in $CTF_2$ and $Pa_{Y_2}$ denote the  parents of variables in $Y_2$. 
    The NC of the distribution encoded by $CTF_2$ ($Z_2$) can be computed as follows.
    \begin{align}\label{eq:ncctf2.1}
        Z_2 &= \sum\limits_{X_2} \frac{\prod_{C'\in CTF_{1,a}}\beta(C')}{\prod_{SP'\in CTF_{1,a}}\mu(SP')} \prod\limits_{y\in Y_2} P(y|Pa_y) \nonumber\\
        &= \sum\limits_{X_2} \textcolor{black}{Q_{1,a}}(X_{1,a},e_1) P(Y_2,e_2~|~Pa_{Y_2}) \text{~~~~~(using Equation~\ref{eq:appD})} 
    \end{align} 
    where $e_2$ are evidence states in $Y_2$.
 \textcolor{black}{Since $X_2=X_{1,a}\cup \{Y_2\setminus E_2\}$ and parent variables in $Pa_{Y_2}$ are present either in $X_{1,a}$ or $Y_2$, the above equation can be re-written as follows.}
 \begin{equation*}
 \color{black}
     Z_2=\sum\limits_{X_2} Q_2(X_2,e_1,e_2) = Q(e_1,e_2)
 \end{equation*}
 \textcolor{black}{Therefore, the NC of $CTF_2$ is an estimate of probability of evidence states $e_1$ and $e_2$.}

    A similar procedure can be repeated for subsequent CTFs to show that the property holds true for all CTFs in the sequence.
\end{proof}

\begin{theorem} \label{thm:post}
Let $I_E$ denote the index of the last CTF in the sequence where the factor corresponding to an evidence variable is added.
  The posterior marginals of variables present in CTFs $\{CTF_{k}, k \geq I_E\}$ are preserved and can be computed from any of these CTFs.
\end{theorem}

\begin{proof}
\color{black}
     Let $\varepsilon_{I_E}=\{e_1,\hdots,e_{I_E}\}$ be the set of all evidence states.
    Let $v$ be a variable present in cliques $C_v\in CTF_{I_E}$, $C'_v\in CTF_{I_E,a}$ and $\tilde{C_v}\in CTF_{I_E+1}$ and let $\beta_N(C_v)$, $\beta_N(C'_v)$ and $\beta_N(\tilde{C_v})$ be the corresponding normalized clique beliefs.  From Proposition~\ref{prop:appBeliefs}, the unnormalized belief of variable $v$ in $C_v$ is same as that in $C_v'$. Therefore, the normalized posterior marginal of $v$ obtained from $C_v$ (denoted as $Q_{I_E}(v|\varepsilon_{I_E}))$) is the same as that obtained from $C_v'$ , as given below.
    \begin{align}\label{eq:4}
    \color{black}
     Q_{I_E}(v|\varepsilon_{I_E}) = \sum \limits_{C_v\setminus v} \beta_N(C_v) = \sum\limits_{C'_v\setminus v} \beta_N(C'_v)
    \end{align}
    Since $CTF_{I_E,a}$ is calibrated (Proposition~6 in \citet{IBIAPR}) and $CTF_{I_E+1}$ is obtained by adding CPDs of variables in $Y_{I_E+1}$ to $CTF_{I_E,a}$, the NC of $CTF_{I_E+1}$ can be computed \textcolor{black}{by summing over all non-evidence variables} as follows.
    \begin{align*}
   Z_{I_E+1} 
         &= \sum\limits_{X_{I_E,a}} \frac{\prod_{C'\in CTF_{I_E,a}}\beta(C')}{\prod_{SP'\in CTF_{I_E,a}}\mu(SP')} \sum\limits_{\textcolor{black}{Y_{I_E+1}\setminus E_{I_E+1}}} P(Y_{I_E+1},e_{I_E+1}|Pa_{Y_{I_E+1}}) \\
         &= \sum\limits_{X_{I_E,a}} \frac{\prod_{C'\in CTF_{I_E,a}}\beta(C')}{\prod_{SP'\in CTF_{I_E,a}}\mu(SP')}   ~~~~~~(\because E_{I_E+1}=\varnothing,~\sum\limits_{Y_{I_E+1}} P(Y_{I_E+1}~|~Pa_{Y_{I_E+1}})=1) \\
         &= Z_{I_E} ~~~~~~~\text{(using Proposition~9 in \citet{IBIAPR})}
    \end{align*}
    Therefore, the posterior marginal of $v$ in $CTF_{I_E+1}$ (denoted as $Q_{I_E+1}(v|\varepsilon_{I_E})$) can be computed from the clique belief of $\tilde{C_v}$  as follows.
    \begin{align*}
        Q_{I_E+1}(v|\varepsilon_{I_E}) &= \sum\limits_{\tilde{C}_v\setminus v} \beta_N(\tilde{C}_v)   \\
        &= \sum\limits_{X_{I_E,a} \setminus v} \frac{1}{Z_{I_E}}  \frac{\prod_{C'\in CTF_{I_E,a}}\beta(C')}{\prod_{SP'\in CTF_{I_E,a}}\mu(SP')} \sum\limits_{Y_{I_E+1}\setminus E_{I_E+1}} P(Y_{I_E+1}, e_{I_E+1}~|~Pa_{Y_{I_E+1}})  \nonumber \\
        &= \sum_{C_v'\setminus v}\beta_N(C_v') ~~~~~~(\because C_v'\in CTF_{I_E,a} ~and~  E_{I_E+1}=\varnothing)\\
        &= Q_{I_E}(v|\varepsilon_{I_E}) ~~~~~~~~~~~~\text{(using Equation~\ref{eq:4})}
    \end{align*}
     The above procedure can be repeated to show that the posterior marginal of $v$ is also consistent in all subsequent CTFs that contain $v$.
\end{proof}